\documentclass[english]{wlscirep}
\usepackage[utf8]{inputenc}
\usepackage[T1]{fontenc}
\usepackage{babel}
\usepackage{amsmath}
\usepackage{graphicx}
\usepackage{fancyhdr}
\usepackage{url}
\usepackage{xcolor}
\usepackage{multicol}
\usepackage{tcolorbox}
\usepackage{hyperref} 
\usepackage{multirow}

\newcommand{\etal}{\textit{et al.} }

\title{A completely annotated whole slide image dataset of canine breast cancer to aid human breast cancer research}

\author[1,+,*]{Marc Aubreville}
\author[2,*]{Christof~A.~Bertram}
\author[3]{Taryn~A.~Donovan}
\author[1]{Christian~Marzahl}
\author[1]{Andreas~Maier}
\author[2]{Robert~Klopfleisch}

\affil[1]{Pattern Recognition Lab, Computer Science \\Friedrich-Alexander-Universit{\"a}t Erlangen-N{\"u}rnberg, Germany}
\affil[2]{Institute of Veterinary Pathology, Freie Universit{\"a}t Berlin, Germany}
\affil[3]{Department of Anatomic Pathology, Animal Medical Center, New York, USA}
\affil[*]{equal contribution}
\affil[+]{corresponding author}

\begin{abstract}
Canine mammary carcinoma (CMC) has been used as a model to investigate the pathogenesis of human breast cancer and the same grading scheme is commonly used to assess tumor malignancy in both. One key component of this grading scheme is the density of mitotic figures (MF). Current publicly available datasets on human breast cancer only provide annotations for small subsets of whole slide images (WSIs). We present a novel dataset of 21 WSIs of CMC completely annotated for MF. For this, a pathologist screened all WSIs for potential MF and structures with a similar appearance. A second expert blindly assigned labels, and for non-matching labels, a third expert assigned the final labels. Additionally, we used machine learning to identify previously undetected MF. Finally, we performed representation learning and two-dimensional projection to further increase the consistency of the annotations. Our dataset consists of 13,907 MF and 36,379 hard negatives. We achieved a mean F1-score of 0.791 on the test set and of up to 0.696 on a human breast cancer dataset.
\end{abstract}

\begin{document}

\flushbottom
\maketitle
\begin{tcolorbox}
The peer-reviewed and published version of this article is available as:

Aubreville, M., Bertram, C.A., Donovan, T.A. et al. A completely annotated whole slide image dataset of canine breast cancer to aid human breast cancer research. Sci Data 7, 417 (2020).

\url{https://doi.org/10.1038/s41597-020-00756-z}
\end{tcolorbox}

\thispagestyle{empty}

\section*{Background \& Summary}
Histologic assessment of tissue is the gold standard in tumor diagnosis and prognostication and is a key component in the selection of the best suited therapy. For the diagnosis of mammary carcinoma, pathologists grade the tumor according to the scheme by Elston and Ellis \cite{Elston:1991dl}. The scheme is based on three criteria: nuclear pleomorphism, tubule formation, and mitotic count. Out of these, a component that is known to have high inter- and intra-rater discordance is the mitotic count, i.e. the relative density of cells undergoing cell division per standardized area \cite{Meyer:2005cl,Meyer:2009eu,Malon:2012je,Bertram:2019vp}. 

While disagreement between experts for individual mitotic figures may be one cause of this, the region of the microscope slide upon which the mitotic count is performed may also have a strong influence, due to patchy distribution of mitotic figures (tumor heterogeneity). Common to most grading schemes, the mitotic count is recommended to be performed in ten consecutive fields representing the field of view area of an optical microscope at 400x magnification (so-called high power field or HPF). The location within the tumor is specified less precisely, but many grading schemes suggest an area at the periphery of the tumor, where the tumor cells are assumed to have greater capacity for proliferation (invasion front). This underlying assumption, however, has not yet been shown to be generally true in mammary carcinoma to the best of the authors' knowledge. In fact,  areas with the highest proliferation density ("hot spots") have been shown to have greater prognostic information than the periphery (invasive edges) based on another quantitative parameter of tumor cell proliferation (Ki67) \cite{staalhammar2018digital}. 

Variable density of mitotic figures within tumors results in less than optimal reproducibility of the MC, which causes a dilemma for its use in prognostication. While we can expect that counting of mitotic figures is of high value due to its representation of tumor biological behavior and growth, due to the aforementioned problems, sub-optimal mitotic counts may lead to inaccurate grading, which can impede prognostication and lead to unexpected or unpredictable outcomes. An optimal mitotic count would thus require a substantially reduced subjective component with precise categorization of individual mitotic figure candidates.  Additionally, the optimal mitotic count would be available for the complete whole slide image (WSI) - or, better: for several WSIs, ideally representing the complete tumor. Neither of these processes can be performed manually by a pathologist within the scope of clinical practice, thus algorithmic support for pathologists by means of a decision support system would be beneficial. This creates a very favorable case for algorithmic support of the pathologist by means of a decision support system. Additionally, since the advent of deep learning-based pattern recognition pipelines, we have seen vast improvements in accuracy of detection systems, given that a sufficient amount of high quality labeled data is available for the task. 

Motivated by this, a number of challenges have been held for the task of mitotic figure detection in recent years  \cite{Roux:2013kn,Roux:2014tt,Veta:2015bi,veta2019predicting}. Leaving aside the results on the MITOS 2012 dataset \cite{Roux:2013kn} (which should no longer be considered state-of-the-art due to selection of the training and test datasets from the same images) the results achieved in these challenges yielded F1 scores of below 0.66 \cite{veta2019predicting}, which can still be regarded as insufficient for clinical use. Multiple factors can be expected to play a role in this: First, we can assume a less than optimal label quality.
Detection of mitotic figures occurs intermittently, thus the probability of missing a portion of mitotic figures on large images is high.
Second, as shown by other authors \cite{Malon:2012je,tsuda2000evaluation,Meyer:2005cl}, agreement on individual, identified mitotic figure candidates, is also far less than perfect (likely depending on the phase in the cell division cycle). Looking only at the final results it is thus hard to judge whether the algorithmic solution is non-optimal or the label noise on the test set is just too high. A third root cause for the results not matching clinical expectations is the low quantity of images and annotations, given the high variability of tissues. Besides the process of mitosis itself, there are numerous other causes for high variability in hematoxylin and eosin (H\&E)-stained microscopy images, including thickness of specimen, concentration and protocol of dying, and optical and calibration properties of the microscope or whole slide scanner that was used.

The answer to all of these challenges can, from our point of view, only be a significant increase in data quantity and quality. All previous datasets of mammary carcinomas only provide annotations for a rather small part of each WSI selected by one expert pathologist \cite{Veta:2015bi}. If algorithmic pipelines should be able to process complete WSIs, this does, however, assume generalization of these areas to the complete slide. Motivated by the findings in our previous work \cite{bertram2019large}, we need to challenge this assumption.  In necrotic or unpreserved regions of the specimen, there may be numerous cells or structures which represent artifacts, but have morphologic overlap with mitotic figures (mitotic-like figures or hard negatives). In order to perform and assess the detection on whole slide images, we thus depend on the availability of annotation data for complete slides. The image quality on the slide is not equally perfect in all regions, which poses an additional burden on the annotation. It is, however, of utmost importance to not only include easily identifiable positive and negative examples, but also to include hard examples. 

Mammary carcinoma is not only prevalent in women, but is also a frequently diagnosed tumor for female dogs (bitches). Due to similarities in epidemiology, biology, and clinical pathology, dogs have been proposed as a model animal to study invasive mammary carcinoma \cite{pinho2012canine,nguyen2018canine}. Besides the application of increased accuracy and cost effectiveness in the treatment of canine tumors, human mammary carcinoma assessment can also benefit from these canine histopathology datasets which provide annotations for entire WSIs, unlike previous datasets of human breast cancer.  

In this work, we present a novel, large-scale dataset of canine mammary carcinoma, providing annotations for 21 complete whole slide images of H\&E-stained tissue. We evaluated the quality of the dataset by using state-of-the-art pattern recognition pipelines based on two stages of deep convolutional neural networks. Additionally, we tested the pipelines trained with our dataset on the largest available human mammary carcinoma dataset (TUPAC16~\cite{veta2019predicting}).

\begin{figure*}[tbp]
	\centering
	\includegraphics[width=0.8\textwidth]{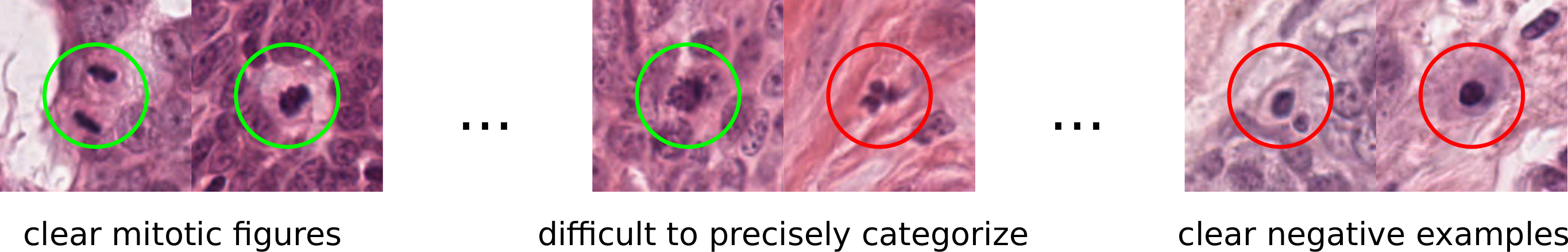}
	\caption{Examples of mitotic figures and structures with a similar appearance. Due to ambiguities, precise classification for some candidates is not straightforward.}
	\label{examples}
\end{figure*}

\section*{Methods}

\subsection*{Selection and preparation of specimen}
All specimens were taken retrospectively from the histopathology archive of an author (R.K.) with approval by the local governmental authorities (State Office of Health and Social Affairs of Berlin, approval ID: StN 011/20). Specimens of breast tissue from bitches had been surgically removed by the treating veterinarian for purely diagnostic purposes for cases suspicious of mammary neoplasia. The tissue had been routinely fixed in formalin and embedded in paraffin. For this study new tissue sections were produced from the tissue blocks and staining with hematoxylin and eosin using an automated slide stainer (ST5010 Autostainer XL, Leica, Germany). Case selection was random and specimens with acceptable tissue quality were included. All images were digitized using a linear whole slide scanner (Aperio ScanScope CS2, Leica, Germany) at a resolution of 0.25 microns per pixel (400X).

\subsection*{Manually Expert Labeled (MEL) Dataset}
Labeling mitotic figures accurately requires a great deal of expertise in the field of tumor pathology. Additionally, it is a labor-intensive task that requires a high level of concentration throughout the process. To set a baseline, an expert pathologist with some years of experience in mitotic figure detection (C.A.B.) screened each WSI twice for mitotic figures. For this, a specialized software solution \cite{sliderunner} was used that provides a screening mode. This mode presents overlapping image patches selected from the WSI in regions where tissue is present. Whenever the expert was done with a given image section, the program would propose the next suitable image patch. This ensures that no portion of the image was left out in this assessment. Besides mitotic figures, the expert annotated similar appearing objects that could be confused with mitotic figures based on their visual appearance, but, do not represent cells in the state of cell division. This was the precondition for the next step: It is widely known, that inter-rater discordance can be high for mitotic figures \cite{Meyer:2005cl,Meyer:2009eu,Malon:2012je,Bertram:2019vp}. To reduce the subjective effect of this rater, we asked a second expert (R.K.), who is a senior pathology expert with several years of experience in mitotic figure assessment. The expert was given the task to assess for each of the annotated objects (the labels were blinded for him) a new label (mitotic figure or mitotic figure look-alike). This results in two independent expert opinions for every single object of interest. For disagreed labels (N = 1,268 / 39,868), agreement was initially obtained by consensus of the first and second pathologist. This preliminary dataset was used for augmented dataset development using machine learning and data analysis techniques (see below).  Final agreement for ground truth, which was used for technical validation of the dataset, was obtained through majority vote by a third pathologist (T.A.D.). Agreement through majority vote was commonly performed in human mitotic figure breast cancer datasets \cite{Roux:2014tt, veta2019predicting}.

\subsection*{Object-Detection Augmented and Expert Labeled (ODAEL) Dataset}
\begin{figure*}[tbp]
	\centering
	\includegraphics[width=0.8\textwidth]{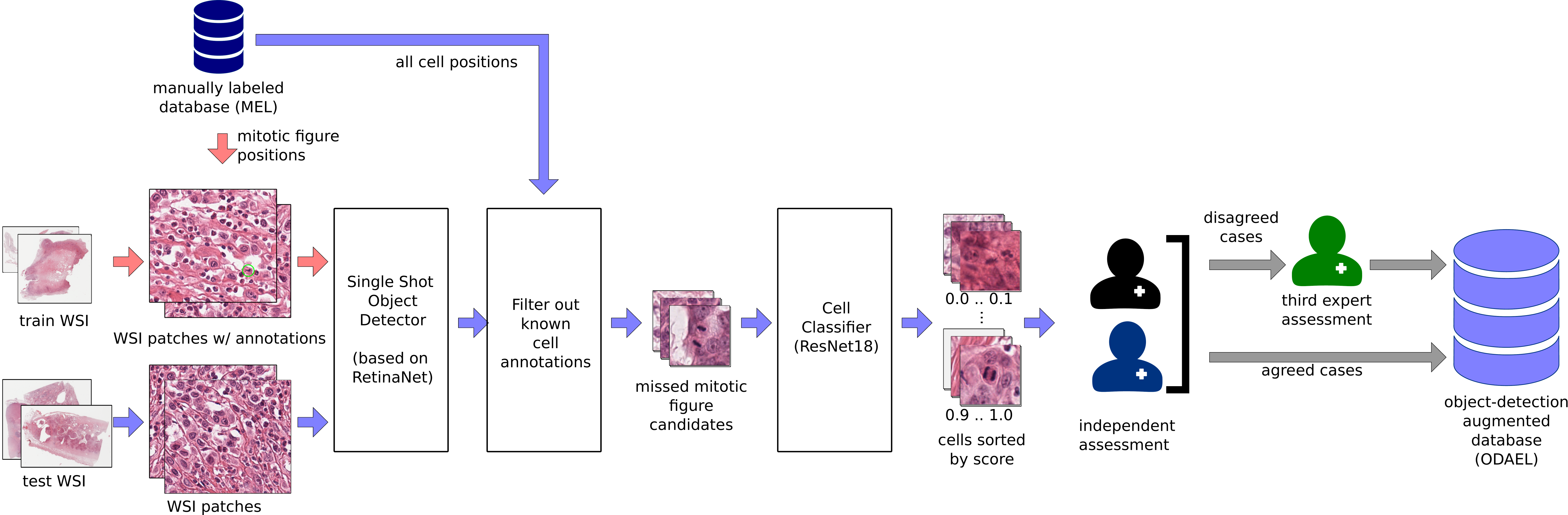}
	\caption{Generation of the object detection-augmented and expert-labeled dataset (ODAEL). Adapted from Bertram \etal  \cite{bertram2019large}.}
	\label{fig:odael}
\end{figure*}

While one expert screened the WSIs with the help of a dedicated software solution and with great diligence, we still have to assume that, due to the partially infrequent occurrence of mitotic figures, the expert might have missed a certain percentage of mitotic figures. To take this into account, we employed a machine learning-based pipeline to find candidates for missed mitotic figures to be presented to the experts (see Figure~\ref{fig:odael}). For this, we first split a preliminary dataset into three parts (each 7 WSIs), which were then subsequently each used once as a test set for a CNN-based object detector, with the WSIs of the other two parts used as training set. We used a customized version \cite{bertram2019large} of RetinaNet \cite{Lin:2017de} for this purpose. All objects detected by the network were subsequently checked for existence in the MEL dataset variant. Second, we trained a cell classifier on 128\,px patches cropped around the cells annotated in the currently used training part of the MEL dataset and ran the inference with all newly identified mitotic figure candidates. The cell patches were grouped into 10 groups according to their model score (where 1.0 represented cells being very likely a mitotic figure, and 0.0 very unlikely, respectively). All mitotic figures were initially shown to the first pathologist and the second pathologist independently, and in case of disagreement, the third expert rendered the final vote, like in the previous dataset variant. Through this procedure, the number of mitotic figures was increased by $6.06\%$ compared to the MEL dataset (see~Table\ref{tab:overview}), which is in line with previous findings \cite{bertram2019large}. The number of non-mitotic cells was increased much more significantly ($+36.22\%$). The reason for this is that all candidates that were identified by the RetinaNet (even those with low scores) were added to the list of non-mitotic figures.
Comparing the individual tumor cases, we can see that the relative increase in mitotic figures is higher for the tumors with less overall mitotic figures. This highlights that missing mitotic figures is more likely for lower grade tumors due to the rareness of the event.

\subsection*{Clustering and Object-Detection Augmented and Expert Labeled (CODAEL) Dataset}
While the previous, machine-learning driven approach was aimed at identifying previously missed mitotic figures, we still have to take into account that the dataset suffers from a certain degree of inconsistency due to misclassification. While it is commonly easy to identify a clear mitotic figure and a clearly non-mitotic cell, there are a significant amount of cells where this differentiation is not easy to make~(see~Figure~\ref{examples}). In this assessment, the experts have to create a visual and perceptual cutoff that pertains to judging what represents a mitotic figure. This cutoff may, however, not be constant over time. 
Both sources of inconsistency, i.e., the varying cutoffs over time as well as plain human error, can be counteracted by clustering with subsequent reevaluation. 

\begin{figure*}[tbp]
	\centering
	\includegraphics[width=0.9\textwidth]{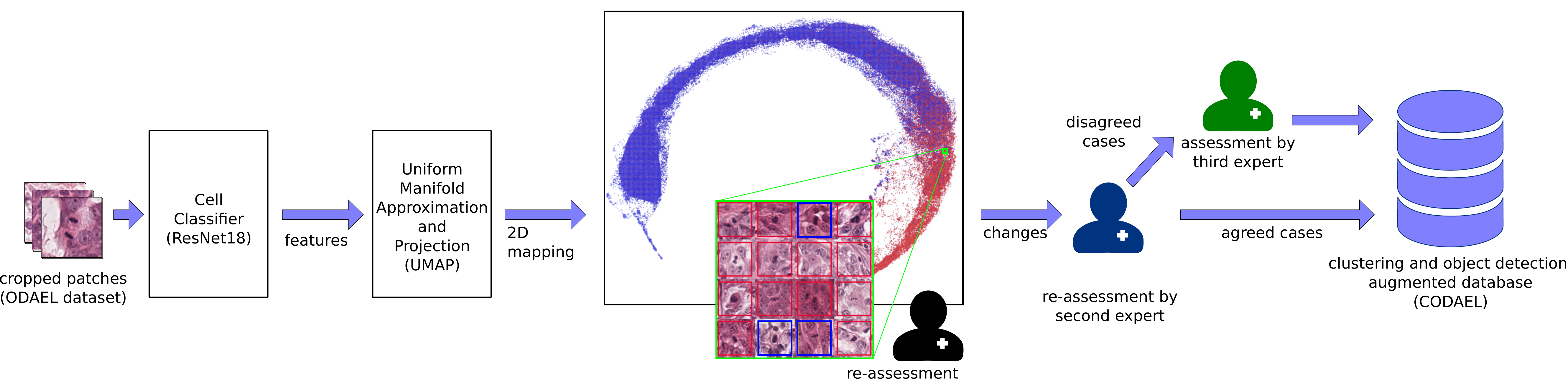}
	\caption{Generation of the Clustering and Object Detection-Augmented Expert Labeled (CODAEL) dataset variant by reassessment of mitotic figures (red) and hard negatives (blue) in a clustered visual representation.}
	\label{fig:clustering}
\end{figure*}

For this, we cropped out patches of all cells (mitotic figures and hard negatives) in the database and trained a ResNet18\cite{He:2016ib}-based classifier for 10 epochs using standard binary cross-entropy. In order to enable a better clustering, we used methods from representation learning in the next step: we removed the last layer of the network and trained the network for another 10 epochs using contrastive loss\cite{hadsell2006dimensionality}. The resulting feature vector of all images were subsequently transformed into a two dimensional representation using uniform manifold approximation and projection (UMAP), resulting in two coordinates in the 2D representation for each image (see~\ref{fig:clustering}).

Next, the image patches were inserted into a new image according to an upscaled version of these coordinates, as described by Marzahl et al\cite{arxiv_EXACT}: To avoid interference of image patches, a grid with tile size according to the patch size was constructed. The patch was then pasted into the grid according to a least distance between grid coordinates and coordinates as given by the projection. Whenever a position in the grid was already filled, the tile was placed on the next possible grid position with least distance. This resulted in an $80,000\times60,000$\,px image, containing all $50,286$ individual cell patches~(see center of Figure~\ref{fig:clustering}). In this image, all patches representing a mitotic figure were given a red box as a rectangle, and all patches representing a non-mitotic cell a blue rectangle. Next, the first pathology expert re-assessed all mitotic figures in this graphical representation. While potentially also introducing a bias, the representation facilitated identification of labeling errors as well as comparisons to similar appearing objects, thus increasing consistency.  Expert driven classification changes occurred for 621 non-mitotic figures (changed to mitotic figures) and 771 mitotic figures (to non-mitotic figures) with a total of 1392 classification changes. 

To reduce the bias introduced by the pipeline and its graphical representation, we presented all changes from the first expert to the second expert, however, 2D mapping was omitted and only patches of 128\,px size were assessed. The second expert disagreed only in a minority of cases. In total, decisions were overruled in 109 cases (7.8\%), resulting in 42 cells classified as mitotic figures and 67 cells classified as non-mitotic cells/structures. The disagreements were lastly given to a third expert for a majority ruling, as shown in Figure ~\ref{fig:clustering}. The final CODAEL datset with majority vote by a third pathologists contains 13,907 mitotic figures, as shown in Table~\ref{tab:overview}.  

\subsection*{Code availability}
All code used in the experiments described in the manuscript was written in Python 3 and is available through our GitHub repository (\url{https://github.com/DeepPathology/MITOS_WSI_CMC/}).
We provide all necessary libraries as well as Jupyter Notebooks allowing tracing of our results.  The code is based on fast.ai \cite{howard2020fastai} and OpenSlide \cite{Goode:2013dm} and provides some custom data loaders for use of the dataset.

\section*{Data Records}

The dataset, consisting of 21 anonymized WSIs in Aperio SVS file format, is publicly available on figshare. Alongside, we provide cell annotations according to both classes in a SQLite3 database. 
For each annotation, this database provides:
\begin{itemize}
    \item The WSI of the annotation
    \item The absolute center coordinates (x,y) on the WSI
    \item The class labels, assigned by all experts and the final agreed class. Each annotation label is included in this, resulting in at least two labels (in the case of initial agreement and no further modifications), one by each expert. The unique numeric identifier of each label furthermore represents the order in which the labels were added to the database.
\end{itemize}

We also provide polygon annotations for the tumor area within the WSI. The polygon annotations consist of multiple coordinates linked to a single annotation. The publicly available libraries provided by the SlideRunner\cite{sliderunner} python package can be used to conveniently extract those. The WSIs and all annotations can also be viewed using this open source software.

Table~\ref{tab:overview} gives an overview about the database and all its variants, sorted by the number of mitotic figures in the CODAEL dataset variant. The number of mitotic figure look-alike cells is almost three times higher than the number of mitotic figures, and this factor is lower for the manually labeled dataset variants. The table also indicates which WSIs were assigned to the training set and which to the test set. 

\begin{figure}[tbp]
    \centering
    \includegraphics[width=\textwidth]{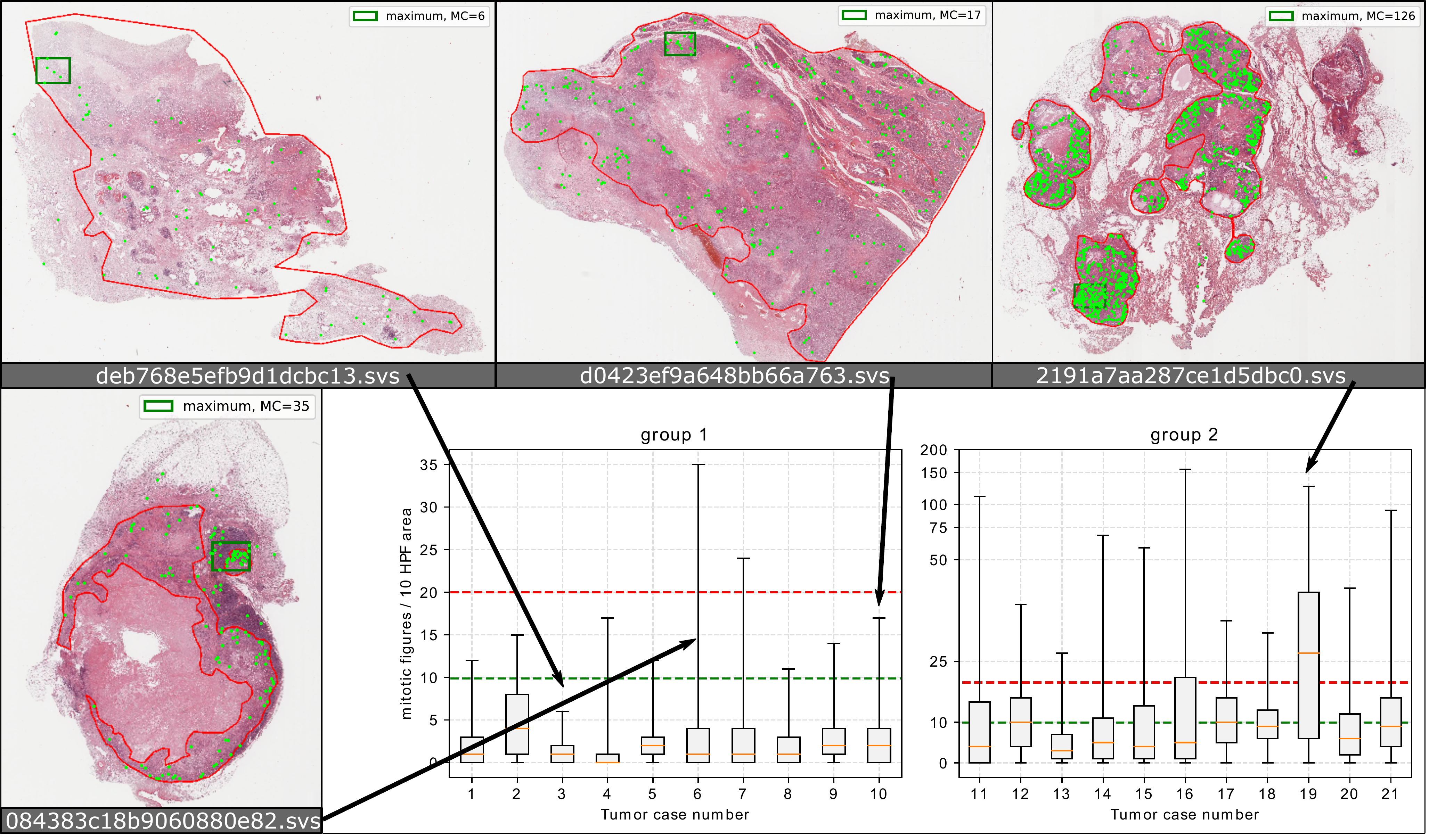}
    \caption{Statistical overview of the count of mitotic figures per area of 10 high power fields (10 HPF, $2.37\,mm^2$) (bottom right). For better visualization, the dataset was split up into two groups (according to the overall sum of mitotic figures). Whiskers indicate absolute maximum, boxes indicate second to third quartile.The dashed red and green lines represent cut-off values. The four images (top row and bottom left) are examples of mitotic figure distribution through the histological section (H\&E stain) using the clustering-aided (CODAEL) dataset variant. Red outlines indicate tumor region. Green dots indicate mitotic figures. The green rectangle in each image indicates the region of maximum mitotic count in an area encompassing 10 HPF (2.37mm2).}
    \label{fig:figure4}
\end{figure}

\begin{table}[tbp]
    \centering
    \resizebox{\textwidth}{!}{
    \begin{tabular}{l|l|r|r|r|l}
    \toprule
         Case No. & File name & tumor area & No. of mitotic figures  & No. of non-mitotic cells  & set  \\
         & &  & (MEL/ODAEL/CODAEL) & (MEL/ODAEL/CODAEL) & \\
         \midrule
1 & 4eee7b944ad5e46c60ce.svs & 66.06\,$\mathrm{mm}^2$ & 47 / 61 / 64 & 114 / 196 / 193 & test \\ 
2 & a8773be388e12df89edd.svs & 37.01\,$\mathrm{mm}^2$ & 64 / 71 / 74 & 204 / 591 / 588 & train \\ 
3 & deb768e5efb9d1dcbc13.svs & 187.43\,$\mathrm{mm}^2$ & 92 / 96 / 84 & 287 / 472 / 484 & train \\ 
4 & e09512d530d933e436d5.svs & 214.97\,$\mathrm{mm}^2$ & 87 / 98 / 102 & 602 / 742 / 738 & test \\ 
5 & 72c93e042d0171a61012.svs & 26.29\,$\mathrm{mm}^2$ & 130 / 151 / 140 & 375 / 680 / 691 & train \\ 
6 & 2d56d1902ca533a5b509.svs & 49.32\,$\mathrm{mm}^2$ & 139 / 155 / 153 & 228 / 365 / 367 & test \\ 
7 & 084383c18b9060880e82.svs & 41.71\,$\mathrm{mm}^2$ & 157 / 173 / 160 & 404 / 547 / 560 & train \\ 
8 & da18e7b9846e9d38034c.svs & 253.10\,$\mathrm{mm}^2$ & 187 / 210 / 211 & 991 / 1,354 / 1,353 & train \\ 
9 & 13528f1921d4f1f15511.svs & 339.93\,$\mathrm{mm}^2$ & 283 / 301 / 292 & 963 / 1,127 / 1,136 & test \\ 
10 & d0423ef9a648bb66a763.svs & 273.88\,$\mathrm{mm}^2$ & 378 / 411 / 354 & 1,143 / 1,596 / 1,653 & train \\ 
11 & 69a02453620ade0edefd.svs & 45.35\,$\mathrm{mm}^2$ & 634 / 642 / 612 & 1,407 / 1,505 / 1,535 & test \\ 
12 & d37ab62158945f22deed.svs & 226.39\,$\mathrm{mm}^2$ & 578 / 651 / 674 & 1,105 / 1,725 / 1,702 & train \\ 
13 & d7a8af121d7d4f3fbf01.svs & 426.92\,$\mathrm{mm}^2$ & 716 / 746 / 720 & 1,832 / 2,373 / 2,399 & train \\ 
14 & 460906c0b1fe17ea5354.svs & 112.24\,$\mathrm{mm}^2$ & 673 / 742 / 754 & 1,199 / 2,480 / 2,468 & train \\ 
15 & b1bdee8e5e3372174619.svs & 231.84\,$\mathrm{mm}^2$ & 812 / 861 / 869 & 1,260 / 1,832 / 1,824 & test \\ 
16 & c4b95da36e32993289cb.svs & 257.01\,$\mathrm{mm}^2$ & 1,097 / 1,114 / 1,085 & 2,454 / 2,944 / 2,973 & train \\ 
17 & 022857018aa597374b6c.svs & 325.81\,$\mathrm{mm}^2$ & 1,290 / 1,344 / 1,320 & 2,463 / 3,106 / 3,130 & test \\ 
18 & 50cf88e9a33df0c0c8f9.svs & 269.25\,$\mathrm{mm}^2$ & 1,197 / 1,339 / 1,337 & 1,632 / 2,550 / 2,552 & train \\ 
19 & 3d3d04eca056556b0b26.svs & 513.28\,$\mathrm{mm}^2$ & 1,383 / 1,465 / 1,447 & 2,110 / 2,933 / 2,951 & train \\ 
20 & 2191a7aa287ce1d5dbc0.svs & 96.38\,$\mathrm{mm}^2$ & 1,449 / 1,485 / 1,462 & 2,155 / 2,609 / 2,632 & train \\ 
21 & fa4959e484beec77543b.svs & 365.91\,$\mathrm{mm}^2$ & 1,949 / 2,035 / 1,993 & 3,598 / 4,408 / 4,450 & train \\
\midrule
& total & 4,360.07\,$\mathrm{mm}^2$ & 13,342 / 14,151 / 13,907 & 26,526 / 36,135 / 36,379& total \\
\bottomrule
\end{tabular}
}
    \caption{Overview of the individual slides of the final dataset. For each slide, the number of mitotic figures and number of non-mitotic structures (hard negatives) is given for each of the three dataset variants: manually expert labeled~(MEL), object detection-augmented and expert labeled~(ODAEL) and clustering and object detection-augmented and expert labeled~(CODAEL).}
    \label{tab:overview}
    
\end{table}
We also investigated the count of mitotic figures. In most grading schemes \cite{Kiupel:2011du,Elston:1991dl} this is defined as ten consecutive high power fields (10 HPF, representing an area of $2.37mm^2$ for the most commonly used microscope settings \cite{Meuten:2016jh}). We chose an area of 10 HPF with an aspect ratio of 4:3 and used a moving window summation over all mitotic figure events to derive the mitotic count. As most grading schemes recommend to perform the mitotic count where there is the highest mitotic activity, we took the absolute maximum of this two-dimensional mitotic count map. The distribution of the mitotic figures per 10 HPF area within the complete tumor area can be seen in Figure~\ref{fig:figure4}: As evident from the examples depicted, there is heterogenous distribution of mitotic figures throughout the tumors. Thus, the mitotic count is highly dependent on the correct determination of the area of maximum mitotic activity. Notably, in almost all cases the higher cutoff value of 10\cite{Elston:1991dl} is exceeded. Since finding the optimal threshold is, however, so strongly dependent on the position, we can assume that these would have to be adjusted for an automatic mitotic figure detection-based grading scheme. In total, the experts annotated a tumor area of $4,360.07mm^2$, greatly exceeding the state-of-the-art in mammary carcinoma datasets, which is given by the TUPAC16 dataset (251.5$\,mm^2$) and similar to the canine cutaneous mast cell tumor dataset~\cite{bertram2019large}.

\subsection*{Getting started}
We provide a github-repository, including all experiments described in this manuscript. The repository includes a jupyter notebook (Setup.ipynb) that will download the dataset from figshare automatically, setting up the environment for all experiments. The repository contains jupyter notebooks using a fast.ai~\cite{howard2020fastai} implementation of RetinaNet. All results used to generate the plots and tables in this paper are provided alongside. Besides the network training notebooks, there is a python script to run inference on the complete dataset, and a script to separate training from test at inference (inference on training is required to properly determine the cutoff thresholds on the WSIs).

In the \texttt{PatchClassifier} subfolder, we provided the implementation of the mitotic figure / non-mitotic figure patch classifier, also used as second stage in our experiments. Alongside with a patch extraction script (to create 128px patches centered around the annotated cell from the WSI), there are jupyter notebooks used for the training of the second stages for each dataset variant and an inference script that we used to yield the final results. 

All database variants described in this paper have been placed in the \texttt{databases} folder.

\section*{Technical Validation}
In order to set a baseline for our novel mammary carcinoma dataset, we performed three experiments: First, we performed a cell classification experiment on cropped-out patches centered around the annotated pattern. Second, we conducted an object detection experiment with the complete WSIs. Lastly, to investigate a domain transfer from canine tissue to human tissue, we repeated this same experiment on the largest publicly available dataset from human mammary carcinoma, TUPAC16 \cite{veta2019predicting}, and subparts thereof. We repeated all experiments five times independently to be able to report mean and standard deviation of the metrics. 

\subsection*{Classification of Centered Patches}
 For the cell classification experiment, we utilized a standard CNN classification pipeline based on a ResNet-18 \cite{He:2016ib} stem, pre-trained on ImageNet \cite{Russakovsky:2015bu}. We used the implementation available in fast.ai~\cite{howard2020fastai} for this and employed the standard image transforms with their default parameters. Due to the high class imbalance, we used minority class oversampling and image augmentation. We initially trained only the randomly initialized classification head for a single epoch, while the rest of the network was frozen. Subsequently, we unfroze the network stem and trained the network for 10 epochs using the cyclic hyper-convergence learning-rate scheme of Smith\cite{Smith:2019jo}. During this, we already employed an early stopping paradigm to prevent overfitting. In this, the model with the highest accuracy score was chosen. As shown in Table~\ref{tab:classification_experiment}, we find a steady increase of the area under the ROC curve (ROC AUC) with each curation step of the dataset: While the manually labeled dataset has a mean ROC AUC value of $0.926$, the clustering- and object-detection supported set already results in a mean value of $0.944$.  

\begin{table}[tbp]
    \centering
    \begin{tabular}{l|r|r|r}
    \toprule
         Metric & MEL variant & ODAEL variant & CODAEL variant \\
         \midrule
Precision & 0.675 $\pm$ 0.055& 0.642 $\pm$ 0.040& 0.679 $\pm$ 0.044\\
Recall & 0.898 $\pm$ 0.023& 0.883 $\pm$ 0.031& 0.891 $\pm$ 0.028\\
ROC AUC & 0.926 $\pm$ 0.014& 0.930 $\pm$ 0.002& 0.944 $\pm$ 0.007\\
         \bottomrule
    \end{tabular}
    \caption{Cell classification experiment, based on cropped-out patches of the manually expert labeled (MEL), object detection-augmented and expert labeled (ODAEL) and the clustering+object detection-augmented and expert labeled (CODAEL) dataset variants. ROC AUC indicates the area under the receiver operating characteristic curve. Values represent mean $\pm$ standard deviation of five independent training and inference runs.}
    \label{tab:classification_experiment}
\end{table}

\subsection*{Detection of Mitotic Figures on WSIs}

For the detection of mitotic figures on whole slide images (WSIs), we employed two state-of-the-art methods: As the primary stage, we used a customized \cite{marzahl2019deep} RetinaNet \cite{Lin:2017de} approach. We used RetinaNet, since it represents a good performance to complexity trade-off, and is available for many machine learning frameworks. As the second stage, we assessed candidates with the patch classifier described in the previous section. We chose this dual stage setup over approaches like Faster RCNN (which integrate two stages into a single network) because this approach offers a higher flexibility for sampling during training, and has been shown to be successful in mitotic figure detection \cite{Li:2018ce}. For the RetinaNet, we used only a single aspect ratio (since mitotic figures and mitotic figure look-alikes can loosely be approximated by a quadratic bounding box), and three scales. The network was based on a ResNet-18\cite{He:2016ib} stem, pre-trained on ImageNet\cite{Russakovsky:2015bu}. Besides standard augmentation methods, we took random crops from the whole slide images based on a sampling scheme~\cite{bertram2019large} that allows it to drive network convergence by selecting patches with true mitotic figures as well as hard negative examples. Since the WSIs already show a quite high diversity in staining, no further color augmentation was used. 

In the first step, the pre-trained network stem was frozen (learning rate = 0), to encourage a fast adaptation of the randomly initialized object detection head. Under these conditions, the network was trained using focal loss as loss function and Adam as optimizer for one epoch (consisting of 5,000 images, randomly selected according to the sampling scheme). Note that we use the term \textit{epoch} in this context differently: as random crops from vastly big images introduce a great deal of randomness, it is complex to assess which parts have been used within the training already - which renders the usual definition of epoch useless. Thus, we arbitrarily define an epoch to contain 5,000 randomly sampled images. As the next step, the network stem was unfrozen and all weights were adapted for two cycles of 10 epochs using the super-convergence scheme \cite{Smith:2019jo}. Next, we trained the network for another 30 epochs, and performed a selection of the model with the lowest validation loss. Towards the end of these 30 epochs, the model would commonly be on the verge of overfitting, thus the model selection process was required to aid generalization. The train- and validation split was performed on the upper (training) vs. lower (validation) part of the WSIs of the training set. While this does not represent a truly statistically independent sample, it was the best compromise to ensure training stability while still keeping the amount of WSIs that are available for training high.

\begin{table}[tbp]
    \centering
    \begin{tabular}{l|r|r|r}
        \toprule
        Network & MEL & ODAEL & CODAEL \\
        \midrule
Single stage (RetinaNet)  & 0.681 $\pm$ 0.014 & 0.702 $\pm$ 0.023 & 0.735 $\pm$ 0.013\\
Dual stage (RetinaNet + ResNet-18)  & 0.707 $\pm$ 0.013 & 0.785 $\pm$ 0.003 & 0.791 $\pm$ 0.012\\
        \bottomrule
    \end{tabular}
    \caption{Performance assessment (F1 score) for mitotic figure detection on the test set of the three different dataset variants, mean and standard deviation for five independent training and inference runs.}
    \label{tab:performance_OD}
\end{table}

In the scope of detecting mitotic figures on whole slide images, we define the detection of a \textit{true positive} (TP) when the detection and the mitotic figure annotation lie within 25\,px distance of one another. If no detection (with a model score above threshold) is present within this distance, it is counted as a \textit{false negative} (FN). On the contrary, if a detection is found outside the vicinity of an annotation, it is classified as a \textit{false positive} (FP).  Using the respective sum of the aforementioned counts over all slides, we define the $F_1$ score as:
\begin{equation}
    F_1 = \frac{2\mathrm{TP}}{2 \mathrm{TP} + \mathrm{FP} + \mathrm{FN}}
\end{equation}

In the same way, we define precision ($\mathrm{Pr}$) and recall ($\mathrm{Re}$) as:
\begin{equation}
    \mathrm{Pr} = \frac{\mathrm{TP}}{ \mathrm{TP} + \mathrm{FP} }
\end{equation}
\begin{equation}
    \mathrm{Re} = \frac{\mathrm{TP}}{ \mathrm{TP} + \mathrm{FN} }
\end{equation}

Test performance improved when using the dual stage algorithm compared to the single stage algorithm, and higher degrees of algorithmic augmentation for dataset development(MEL to CODEAL dataset variant) further improved the $F_1$ metric (see Table~\ref{tab:performance_OD}). Standard deviations of five independent training and testing runs were small overall, proving that this dataset may be used to train algorithms reproducibly.


\subsection*{Methods of label agreement}
Previous datasets of mitotic figures in human mammary carcinoma used a majority vote by a third pathologist for disagreed labels \cite{Roux:2014tt, veta2019predicting}. Therefore we also used this approach to establish the ground truth for the technical validation study. In a previous canine dataset of mitotic figures, a consensus by the same pathologists was used for ground truth \cite{bertram2019large} and we used this approach for the training of the networks used in the augmented datasets. This section aims to briefly compare these two methods. For the CODEAL datset variant, the consensus contained 30 more mitotic figures, thus there is a negligible difference in the total number of mitotic figures (0.22\%). Training and testing with the consensus variant yielded an $F_1$ score of 0.790, which is comparable to the majority vote variant with the third expert ($F_1$ of 0.791 $\pm$ 0.012). 

Both methods for label agreement have potential biases in regard to label consistency (reproducible decision criteria) and label accuracy (true label class). While the majority vote method may potentially be more representative of the general expert opinion, introducing another expert with variable decision criteria might potentially reduce label consistency. Although future studies need to examine the influence of these two different labeling methods, we could not find a notable difference in the total number of labels or algorithmic performance for our datasets.

\subsection*{Inter-Species Transfer: Applicability on Human Breast Tissue}
Human mammary carcinoma is a cancer of high prevalence worldwide. Due to the similarity of canine mammary carcinoma and human mammary carcinoma, we aimed to investigate how applicable a system trained on the dataset proposed in this work would be on human mammary carcinoma. Between datasets of different origins, we can expect to observe a domain shift, effectively reducing the performance in cross-domain detection results. This domain shift might be caused by biological differences in tumor or normal tissue between humans and dogs and variable tissue processing workflows between different laboratories including variable WSI acquisition. 
Finally, we can also expect a difference in expert opinions between different ground truth datasets which might be reflected in lower recognition performance.

\begin{figure}
    \centering
    \includegraphics[width=0.7\textwidth]{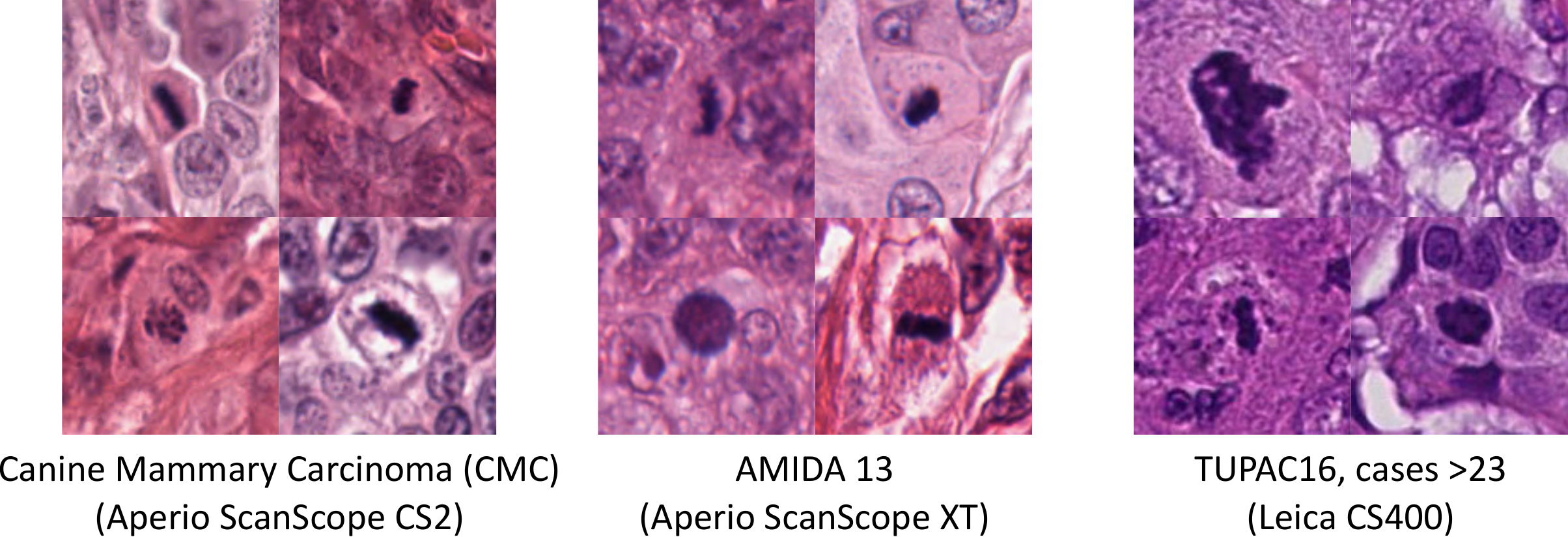}
    \caption{Patches containing mitotic figures from our canine dataset (left), the AMIDA13 cases within TUPAC16 (middle), and the remaining cases of TUPAC16 (right). The clear difference in color representation causes a domain shift.}
    \label{fig:colorvariations}
\end{figure}

To investigate this, let us have a brief look at the TUPAC16 \cite{veta2019predicting} dataset: According to Veta \etal, the dataset was acquired using two scanners of different types at three clinical environments. The first 23 cases, previously released as the MICCAI AMIDA-13 dataset, were acquired using the Aperio ScanScope XT, while the remaining 84 cases were acquired using the Leica CS400 scanner. Both scanners have very different color representation (see Figure \ref{fig:colorvariations}), and thus cause a severe domain shift, that shall, however, not be the scope of this work, as we only wanted to investigate the general transferability of models trained on canine tissue. The images used for the canine specimens (scanned with Aperio ScanScope CS2) seem to have similar color representation as compared to human images scanned with the Aperio ScanScope XT.

To undertake the question of reduced generalization caused by different opinions of the annotators, we re-labeled the TUPAC16 set with very similar methods, using experts from the present study \cite{bertram2020pathologist}.

\begin{table}[tbp]
    \centering
    \begin{tabular}{l|l|c|r|c|l|l}
    \toprule
training conditions &  \multicolumn{4}{c|}{test conditions}  & \multicolumn{2}{c}{$F_1$ Score} \\
& dataset & labels & cases & scanner & single stage & dual stage \\
       \midrule
  only CMC & TUPAC train & \multirow{5}{*}{original labels \cite{veta2019predicting}}  & 73 & L,A & 0.528 $\pm$ 0.029& 0.544 $\pm$ 0.014\\
 only CMC & TUPAC test & & 34 & L  & 0.322 $\pm$ 0.032 & 0.268 $\pm$ 0.039 \\
only CMC & AMIDA train  & & 12 & A &  0.524 $\pm$ 0.022& 0.574 $\pm$ 0.019\\
only CMC &  AMIDA test  & & 11 & A & 0.546 $\pm$ 0.044& 0.579 $\pm$ 0.026\\
TO and MS on AMIDA-train&  AMIDA test  & &  11 & A & 0.584 & 0.628 \\
\midrule
only CMC &  TUPAC-train &  \multirow{6}{*}{re-labeled \cite{bertram2020pathologist}} & 73 & L,A &0.564 $\pm$ 0.038& 0.573 $\pm$ 0.019\\
only CMC & TUPAC-test & & 34 & L &  0.298 $\pm$ 0.044& 0.218 $\pm$ 0.036\\
only CMC &  AMIDA-train  & & 12 & A &  0.592 $\pm$ 0.043& 0.645 $\pm$ 0.016\\
only CMC &  AMIDA-test  & &11 & A & 0.594 $\pm$ 0.047& 0.635 $\pm$ 0.028\\
TO and MS on AMIDA-train &  AMIDA-test  & & 11 & A & 0.628 & 0.696\\
TL on AMIDA-train & AMIDA-test & & 11 & A &  0.720 $\pm$ 0.022& 0.733 $\pm$ 0.007\\
       \bottomrule
    \end{tabular}
    \caption{Mitotic figure detection performance ($F_1$ Score), when trained on the final canine mammary carcinoma (CMC) dataset and tested on the TUPAC16 dataset and its subsets (including AMIDA-13), without any domain adaptation and with only threshold optimization (TO) and model selection (MS), or with transfer learning (TL) performed on the target domain. Values given are mean and standard deviations of five independent training and inference runs. The histological images (cases) were obtained with Aperio ScanScope XT (A) or Leica SCN400 (L) scanner, both with a resolution of 0.25 microns per pixel (400X magnification) }
    \label{tab:transfertask}
\end{table}

Looking at Table~\ref{tab:transfertask}, we can see that the domain transfer task yields mean F1 scores of 0.544 on the training set of TUPAC16. At first glance, this seems like a significant reduction in performance, compared to what the system trained on the same dataset achieved on the test set presented in this work. 

As shown in Table~\ref{tab:transfertask}, the performance is already better if we limit the task to the AMIDA13 dataset, or the test set thereof, which can likely be explained by the similar color representation between the human and canine images.  In contrast, testing on the TUPAC-test set, which comprises exclusively of images from the Leica SCN400 scanner, has a lower $F_1$ score. As the second stage of our algorithm is specifically trained to distinguish difficult patterns, it is likely to show a stronger dependency on the domain.  Thus a change in color representation might lead to inferior performance, compared to the single stage approach as shown in the present study.

Performance was further improved when testing on the re-labeled ground truth, which likely has higher annotation consistency to the canine dataset. This underscores the importance of consistent labeling methods when algorithms are to be tested on independent datasets. We further performed model selection (MS) and optimization of the threshold (TO) on the AMIDA training set as a step towards further domain adaptation, resulting in a higher F1 score of 0.696.    

A common approach when a big and more representative dataset and a smaller target-domain dataset are available is \textit{transfer learning}. For this, a model that was trained on the original, large dataset is fine-tuned on the smaller dataset. In order to estimate the remaining domain shift of the Aperio ScanScope-scanned slides of the TUPAC dataset (i.e., the AMIDA-13 dataset), we thus additionally performed model fine-tuning using the training part of the AMIDA-13 dataset, and evaluated it on the respective test set. Our results in Table \ref{tab:transfertask} indicate that there is a residual domain shift, likely caused by variability in tissue processing or quality, or by the fact that only hot spot regions were annotated for the AMIDA-13 dataset, and not the whole WSI. 

In summary, we can observe a strong domain shift that can be largely attributed to the scanner that was used for the digitization of images, and a much lesser domain shift caused by a combination of processing differences (such as staining, section thickness, etc.) or species. Additionally, we found a decrease in performance that can be attributed to label inconsistency, either caused by differences in expert opinion or by using a different labeling workflow. Results of this domain transfer experiment suggest that this dataset may be valuable for training or testing mitotic figure algorithms for human breast cancer as we provide annotations for entire WSIs.


\section*{Usage Notes}

Annotations are provided in the SlideRunner database format \cite{sliderunner}, which can be also used to view the WSIs with all annotations, but also in the popular MS COCO format. The latter, however, only contains the final annotation class and not the annotation history (i.e., the multiple labels given by multiple experts).  



\section*{Author Contributions}

MA and CAB wrote the manuscript and carried out the main research and analysis tasks of this work. MA carried out data analysis, training of networks and method development. CM provided the code for the clustering experiments and helped in general method development. AM and RK provided guidance for method development and reviewed the manuscript. CAB and RK provided all the annotation data for this dataset. TAD served as a third expert for the annotation and contributed to the manuscript.

\section*{Acknowledgements}

CAB gratefully acknowledges financial support received from the Dres. Jutta \& Georg Bruns-Stiftung für innovative Veterinärmedizin.




\section*{Competing financial interests}

The author(s) declare no competing financial interests.
\bibliography{Literature.bib}

\end{document}